\documentclass[a4paper, 10pt, conference]{IEEEtran}
\IEEEoverridecommandlockouts
\usepackage{cite}
\usepackage{amsmath,amssymb,amsfonts}
\usepackage{multirow}
\usepackage{algorithmic}
\usepackage{graphicx}
\graphicspath{{assets/}}

\ifCLASSOPTIONcompsoc
    \usepackage[caption=false, font=footnotesize, labelfont=sf, textfont=sf]{subfig}
\else
\usepackage[caption=false, font=footnotesize]{subfig}
\fi

\usepackage{textcomp}
\usepackage{hyperref}
\usepackage[nameinlink]{cleveref}
\crefname{figure}{Fig.}{Figs.}
\crefname{table}{Table}{Tables}
\usepackage{booktabs}
\usepackage{multirow}
\usepackage{makecell}
\usepackage{rotating}
\usepackage{xcolor}
\usepackage{tabularx}
\def\BibTeX{{\rm B\kern-.05em{\sc i\kern-.025em b}\kern-.08em
    T\kern-.1667em\lower.7ex\hbox{E}\kern-.125emX}}

\usepackage{tikz}
\usepackage{lipsum}

\newcommand\copyrighttext{%
  \small \centering{PCS 2025, Aachen, Germany, Dec. 2025}  \\ \vspace{1mm}
  \footnotesize \textcopyright 2025 IEEE. Personal use of this material is permitted.
  Permission from IEEE must be obtained for all other uses, in any current or future
  media, including reprinting/republishing this material for advertising or promotional
  purposes, creating new collective works, for resale or redistribution to servers or
  lists, or reuse of any copyrighted component of this work in other works.
  }
\newcommand\copyrightnotice{%
  \begin{tikzpicture}[remember picture,overlay]
    \node[anchor=north,yshift=-8pt] at (current page.north) {%
      \parbox{\dimexpr\textwidth\relax}{\copyrighttext}%
    };
  \end{tikzpicture}%
}

\begin{document}
\bstctlcite{IEEEexample:BSTcontrol}

\title{
How Universal Are SAM2 Features?
}

% % Authors 
\author{
\IEEEauthorblockN{Masoud Khairi Atani\IEEEauthorrefmark{1}, Alon Harell\IEEEauthorrefmark{1}, Hyomin Choi\IEEEauthorrefmark{2}, Runyu Yang\IEEEauthorrefmark{1}, Fabien Racapé\IEEEauthorrefmark{2}, and Ivan V. Baji\'{c}\IEEEauthorrefmark{1} 
}

    % \vspace{5\baselineskip} % Add a little space below authors
    \parbox[t]{0.49\textwidth}{ % Adjust width as needed
        \centering
        % \\
        \IEEEauthorrefmark{1}%\textit{School of Engineering Science} \\
\textit{Simon Fraser University}\\
Burnaby, BC, Canada \\
\{masoud\_khairi\_atani, aharell, runyuy, ibajic\}@sfu.ca}
    \hfill % Pushes the next box to the right
    \parbox[t]{0.41\textwidth}{ % Adjust width as needed
        \centering
        % \\
        \IEEEauthorrefmark{2}\textit{InterDigital AI Lab} \\
Los Altos, CA, USA \\
\{hyomin.choi, fabien.racape\}@interdigital.com}
% \IEEEauthorblockA{
}

\maketitle

\copyrightnotice

\begin{abstract}
The trade-off between general-purpose foundation vision models and their specialized counterparts is critical for efficient feature coding design and is not yet fully understood. We investigate this trade-off by comparing the feature versatility of the general-purpose Hiera encoder against the segmentation-specialized Segment Anything Model 2 (SAM2). Using a lightweight, trainable \emph{neck} to probe the adaptability of their frozen features, we quantify the information-theoretic cost of specialization. Our results reveal that while SAM2's specialization is highly effective for spatially-related tasks like depth estimation, it comes at a cost. The specialized SAM2 encoder underperforms its generalist predecessor, Hiera, on conceptually distant tasks such as pose estimation and image captioning, demonstrating a measurable loss of broader semantic information. A novel \emph{cross-neck} analysis on SAM2 reveals that each level of adaptation creates a further representational bottleneck. Our analysis illuminates these trade-offs in feature universality, providing a quantitative foundation for designing efficient feature coding and adaptation strategies for diverse downstream applications.
 \end{abstract}


\section{Introduction}
\label{sec:introduction}

The field of computer vision has been reshaped by large-scale foundation models, typically built on the Vision Transformer (ViT) architecture \cite{dosovitskiy2021an}. While some models, such as CLIP, DINOv2, and Hiera \cite{radford2021learning, oquab2024dinov, ryali2023hiera}, are designed for general-purpose feature extraction, a new class of powerful, specialized models has emerged. The Segment Anything Model 2 (SAM2) \cite{ravi2025sam} is a prime example, fine-tuned from a Hiera backbone for state-of-the-art segmentation.  

The success of SAM2 in segmentation and related domains like Medical Image Segmentation and Shadow Detection \cite{chen2025samadapter}
naturally leads to its adoption for a wider range of applications. However, this raises fundamental questions: What is the trade-off in feature utility when a general-purpose encoder like Hiera is specialized? Does this process create an informational bottleneck that limits versatility for other tasks? Furthermore, if a model is adapted to a task once, what is the cost of adapting it again? Does information loss compound? Answering these questions is critical for understanding the true potential and limitations of such models, especially in resource-constrained scenarios of feature compression, where a single transmitted bitstream must serve multiple, sometimes unknown, downstream tasks. This is a significant challenge in universal feature coding. The versatile and generalizable capabilities of Segment Anything Models (SAMs) have garnered attention within the MPEG Feature Coding for Machines group as a potential foundation for future work \cite{sam_mpeg_fcm}.

This paper provides a systematic investigation into these questions by directly comparing the frozen Hiera and SAM2 encoders. We introduce a diagnostic framework, centered on a trainable adaptation module which we call a \emph{neck}, to probe the inherent versatility of their features. Our approach allows us not only to assess performance on new tasks but also to quantify, via a novel \emph{cross-neck} analysis on SAM2, the information lost when features are sequentially specialized. We evaluate this methodology across a diverse set of vision tasks to understand the trade-offs of task adaptation. Our contributions are as follows:
\begin{itemize}

\item We propose a neck and novel cross-neck methodology to assess feature versatility and quantify the information cost of model specialization,
providing a framework to evaluate candidate features for universal coding schemes.

\item We find a clear trade-off: specialization boosts performance on related tasks but creates an information bottleneck, degrading performance on dissimilar tasks compared to the original general-purpose encoder.

\item Our cross-neck analysis on SAM2 further demonstrates that sequential adaptation creates compounding representational bottlenecks, measurably degrading feature utility for subsequent tasks.
\item We show that distillation and increased adapter capacity effectively mitigate information loss and reduce the performance gap.
\end{itemize}

\begin{figure}[tb]
\centering
\includegraphics[width=0.7\linewidth]{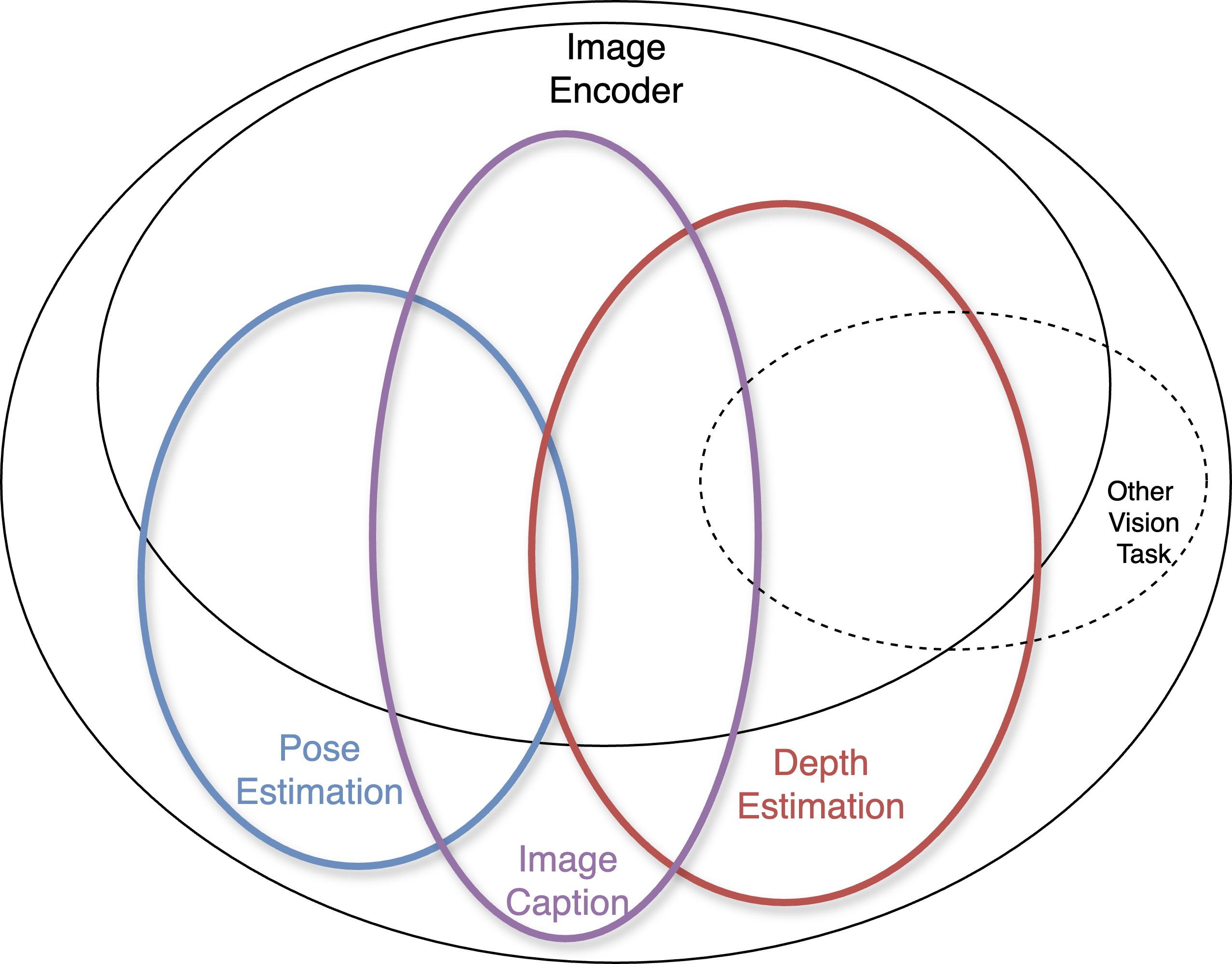}

\caption{Conceptual diagram of the information captured by an image encoder (inner ellipse) relative to total image information (outer ellipse). We quantify the encoder's latent knowledge on diverse vision tasks (colored ellipses).
}
\label{fig:information_content}
\end{figure}

\section{Method}
\label{sec:method}

\begin{figure}[tb]
    \centering
    \includegraphics[width=0.9\linewidth]{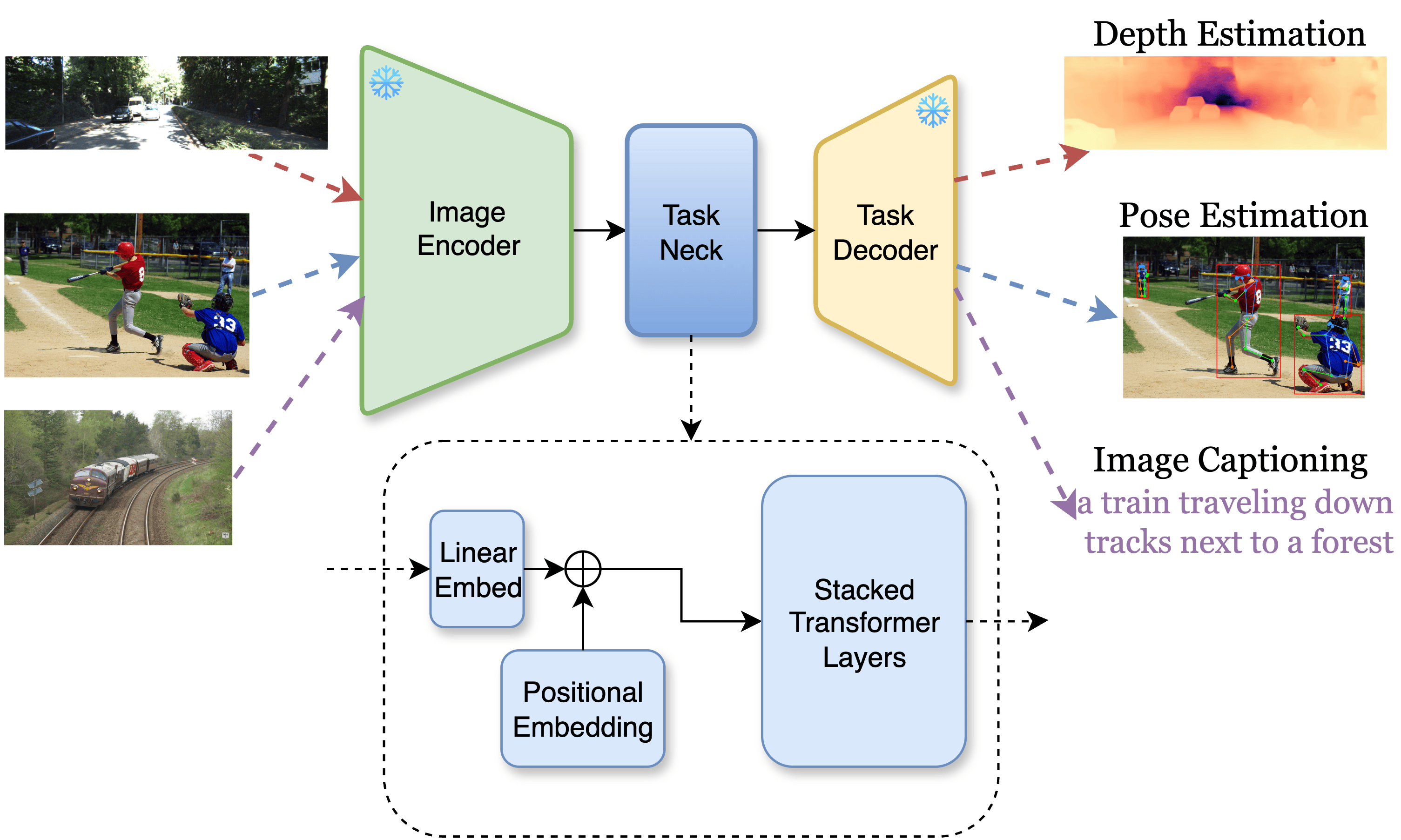}
    
    \caption{A unified model architecture serving as a template for each vision task in single-neck adaptation setting. Our framework uses a frozen encoder, a frozen task decoder (head), and a trainable neck.}
    \label{fig:model_arch_all_tasks}
\end{figure}

Our methodology is designed to quantify the versatility of features (as conceptualized in \cref{fig:information_content}) from frozen image encoders by measuring their adaptability to diverse downstream tasks. Our core framework, shown in \cref{fig:model_arch_all_tasks}, consists of three components: a frozen encoder, a pretrained and frozen task-specific decoder (sometimes known as task “head”), and a lightweight, trainable \emph{neck} network that bridges them. The encoders we investigate are Hiera (\texttt{hiera\_base\_plus}) and SAM2 (\texttt{sam2.1\_hiera\_base\_plus}). By training only the neck, we can isolate and measure the adaptability of the encoder's features.

The neck is a critical component, designed to be powerful enough to bridge the representational gap but lightweight enough to ensure efficient training. We employ a Transformer-based neck to capture long-range dependencies and enable flexible feature transformation, providing a general solution rather than one tailored to a single task. It processes the encoder features with a linear embedding layer, adds a learnable positional encoding, and feeds the result into a stack of transformer layers (\cref{fig:model_arch_all_tasks}). To study the impact of adapter capacity, we define three variants: \texttt{neck\_2L}, \texttt{neck\_4L}, and \texttt{neck\_6L}, configured with 2, 4, and 6 Transformer layers and attention heads, respectively.

We employ this framework on three distinct downstream tasks to comprehensively probe the encoders' capabilities:
monocular depth estimation (using DPT~\cite{oquab2024dinov}), human pose estimation (ViTPose~\cite{xu2022vitpose}), and image captioning (mPLUG~\cite{li2022mplug}). These distinct tasks were chosen to span a range of visual understanding, from dense, spatial prediction to high-level semantic interpretation. Further details on each task configuration are provided in Section~\ref{sec:sub:downstream_tasks}.

We conduct two main experiments using this framework. The first, \textbf{Single-Neck Adaptation} (\cref{fig:model_arch_all_tasks}),
evaluates the versatility of the Hiera and SAM2 encoders. For each task, a dedicated neck is trained to adapt the frozen encoder features for the frozen decoder (head), enabling direct comparison of task-relevant knowledge within each encoder.

\begin{figure}[tb]
    \centering
    \includegraphics[width=0.9\linewidth]{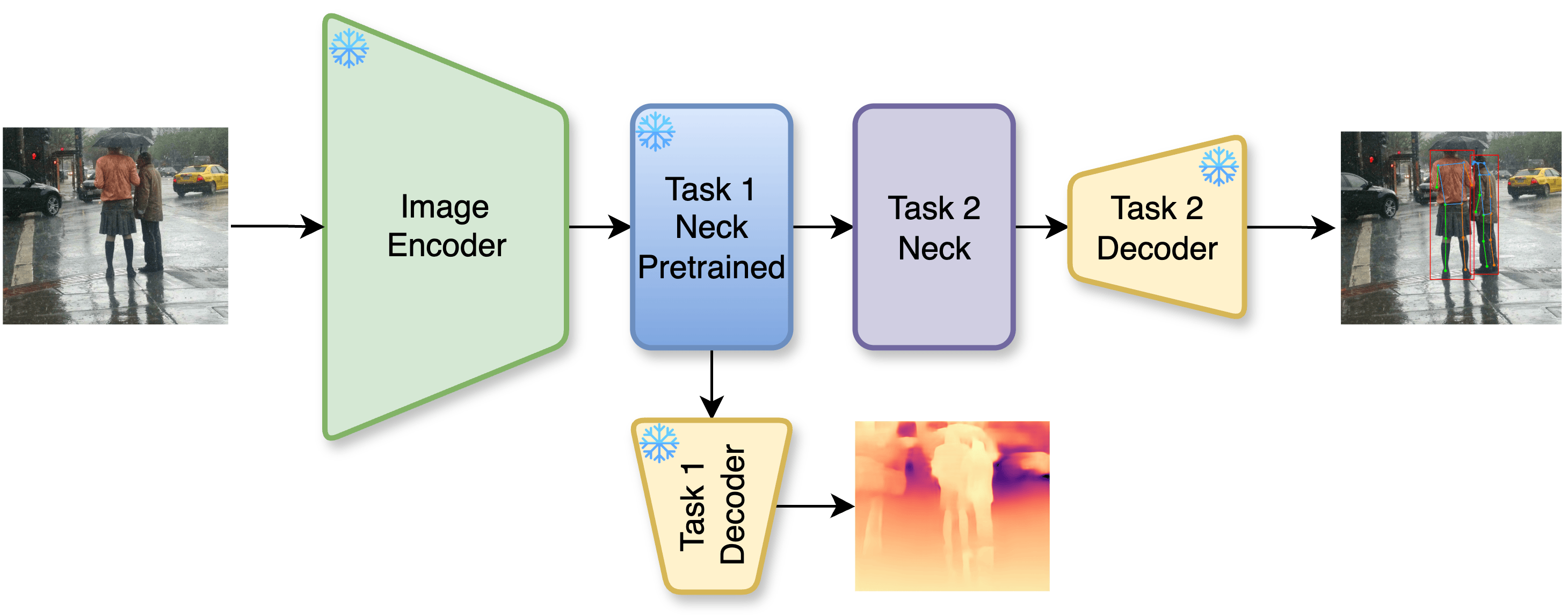}
    \caption{Cross-neck model designed to measure the representational bottleneck of sequential task adaptation by using features from a frozen, pre-trained neck (task 1 neck) to train a second neck (task 2 neck) on a new task.}
    \label{fig:model_architecture_cross_neck}
\end{figure}

The second experiment, a \textbf{Cross-Neck Analysis} (\cref{fig:model_architecture_cross_neck}), is designed to explicitly measure the information bottleneck created by specialization. Here, we first train and freeze a neck for a primary task (e.g., depth estimation). Then, using the output of this first neck as input, we train a second neck on a different target task (e.g., pose estimation). This sequential adaptation directly probes how specializing features for one task degrades their utility for another.

For both setups, we explore two training strategies: one relies solely on the task-specific loss ($L_T$) and another incorporates knowledge distillation via a weighted sum:
\begin{equation}
    L = \alpha L_{D} + (1 - \alpha) L_{T}
\end{equation}
The distillation loss, $L_D$, is the mean squared error (MSE) between the output features of the neck and those of the corresponding baseline encoder. Its influence is controlled with a linear annealing schedule for $\alpha$: it starts at 1, forcing the neck to primarily mimic the expert features, then decays to 0 over the course of training, gradually shifting focus to task loss.

\subsection{Mutual Information Estimation}
Task-relevant information captured by an encoder is quantified by estimating the mutual information (MI) between its adapted features and those from an expert baseline. For each task, we select a high-performing model (e.g., ViTPose for pose estimation) and use its latent features as an expert reference. By measuring the MI against this reference, we can assess how well the Hiera/SAM2 encoder's representations align with the informational demands of downstream applications.

Our methodology involves processing an input image through two parallel pathways. The first pathway produces the task features, $F_{\text{T}}$, by passing the image through the frozen baseline encoder. The second pathway yields our adapted features, $\widehat{F}_{\text{T}}$, by passing the same image through the Hiera/SAM2 encoder and the task-specific neck. The MI between these two feature sets is then computed using an estimator, $\mathcal{I}$:
\begin{equation}
    I(\widehat{F}_{\text{T}}; F_{\text{T}}) \approx \mathcal{I}(\widehat{F}_{\text{T}}; F_{\text{T}})
\end{equation}
This process is repeated for each task, providing a quantitative measure of the information overlap between the adapted features and the downstream task.

\section{Experimental Setup}
\label{sec:experiments}

Our analysis centers on comparing a general-purpose encoder with its specialized descendant. For the generalist model, we use the frozen pretrained \texttt{hiera\_base\_plus} from the official Hiera repository~\cite{ryali2023hiera}. For the segmentation-specialized model, we use the frozen pretrained image encoder of \texttt{sam2.1\_hiera\_base\_plus} variant of the Segment Anything Model 2 (SAM2)~\cite{ravi2025sam}. We detail the datasets, evaluation metrics, and task-specific implementation choices below.

\subsection{Downstream Task Configurations}
\label{sec:sub:downstream_tasks}

For \textbf{monocular depth estimation}, we evaluate on the KITTI dataset~\cite{Geiger2013IJRR} using the standard Eigen split~\cite{eigen2015predicting}. Performance is reported using Absolute Relative error (Abs Rel) and Root Mean Squared Error (RMSE)~\cite{eigen2014depth}. Our task-specific baseline consists of the DINOv2 encoder paired with its corresponding pretrained DPT decoder, a widely recognized, high-performing model for this task.

For \textbf{human pose estimation}, experiments are conducted on the MS COCO keypoint detection dataset~\cite{coco2014}. We report mean Average Precision (AP) and mean Average Recall (AR), calculated based on Object Keypoint Similarity (OKS). Our approach adapts the ViTPose-Base architecture, a strong Transformer-based baseline, pairing our source encoder+neck with its simple convolutional decoder.

Finally, for \textbf{image captioning}, we use the COCO Caption dataset~\cite{chen2015microsoft} (Karpathy split). Caption quality is measured using BLEU@4 (B@4)~\cite{papineni2002bleu} and CIDEr~\cite{vedantam2015cider}. Our model adopts the mPLUG vision-language model, replacing its CLIP visual encoder with our encoder+neck module while keeping the BERT-based text decoder frozen.

\subsection{Information-Theoretic Metrics}

Beyond task-specific performance, we employ a suite of metrics to analyze the feature representations at a deeper level. First, we leverage statistical and divergence-based estimators common in generative modeling, including Fr\'echet Distance (FD)~\cite{heusel2017gans} and KID-like Kernel Distances using both RBF and polynomial kernels (KD\textsubscript{rbf}, KD\textsubscript{poly})~\cite{binkowski2018demystifying}. These are supplemented with baseline measures of cosine similarity and an MI estimate under a simplifying Gaussian assumption (which we call MI 1D Gauss).

To complement these distributional measures, we also employ two neural estimators to directly quantify mutual information (MI) between high-dimensional feature distributions. We use the established Mutual Information Neural Estimator (MINE)~\cite{belghazi2018mutual} and the Latent Mutual Information (LMI) estimator~\cite{gowri2024approximating}, with the latter specifically designed for high-dimensional settings by first learning low-dimensional projections of its inputs. This diverse suite of metrics provides a holistic view, with FD and KD measuring overall distributional similarity while the neural estimators provide a more direct assessment of shared information.
% \vspace{-0.3cm}
\section{Results}
\label{sec:result}

\begin{table}[tb]
\caption{Performance of baseline vs. adapted models on three tasks ($\downarrow$: lower is better;$\uparrow$: higher is better). Adapted results are using the \texttt{neck\_6L} with distillation. Results are shown as mean$_{\scriptscriptstyle \pm std}$ over three runs.}
% ($\downarrow$: lower is better;$\uparrow$: higher is better)
\label{tab:performance_summary}
\centering
\footnotesize % Use a smaller font size for a tight fit
\setlength{\tabcolsep}{2pt} % Reduce space between columns
\resizebox{\columnwidth}{!}{%
\begin{tabular}{l|cc|cc|cc}
\toprule
% Headers with line breaks for space
\textbf{Model} & \multicolumn{2}{c}{\textbf{Performance}} & \multicolumn{2}{c}{\makecell{Feature \\ Distance $\downarrow$}} & \multicolumn{2}{c}{\makecell{MI \\ Estimation}} \\
\midrule

% --- KITTI Depth Estimation ---
 \textbf{Depth} & $AbsRel$ $\downarrow$ & $RMSE$ $\downarrow$ & $FD$ & $KD_{rbf}$ & $LMI$ & $MINE$ \\
\cmidrule(r){2-3} \cmidrule(lr){4-5} \cmidrule(l){6-7}
DPT & $7.90$ & $3.09$ & -- & -- & -- & -- \\
Hiera+neck & $8.34_{\scriptscriptstyle \pm .03}$ & $3.21_{\scriptscriptstyle \pm .01}$ & $9.54_{\scriptscriptstyle \pm .04}$ & $9.83_{\scriptscriptstyle \pm .06}$ & $10.35_{\scriptscriptstyle \pm .05}$ & $10.74_{\scriptscriptstyle \pm .04}$ \\
SAM2+neck & $8.04_{\scriptscriptstyle \pm .05}$ & $3.07_{\scriptscriptstyle \pm .01}$ & $9.28_{\scriptscriptstyle \pm .05}$ & $9.60_{\scriptscriptstyle \pm .09}$ & $10.33_{\scriptscriptstyle \pm .08}$ & $10.54_{\scriptscriptstyle \pm .05}$ \\
\midrule

% --- COCO Human Pose Estimation ---
\textbf{Pose} & $AP$ $\uparrow$ & $AR$ $\uparrow$ & $FD$ & $KD_{rbf}$ & $LMI$ & $MINE$ \\
\cmidrule(r){2-3} \cmidrule(lr){4-5} \cmidrule(l){6-7}
ViTPose  & $76.0$ & $78.9$ & -- & -- & -- & -- \\
Hiera+neck & $73.3_{\scriptscriptstyle \pm .2}$ & $76.6_{\scriptscriptstyle \pm .4}$ & $0.353_{\scriptscriptstyle \pm .002}$ & $0.20_{\scriptscriptstyle \pm .01}$ & $8.03_{\scriptscriptstyle \pm .02}$ & $11.63_{\scriptscriptstyle \pm .04}$ \\
SAM2+neck & $69.8_{\scriptscriptstyle \pm .3}$ & $73.2_{\scriptscriptstyle \pm .2}$ & $0.392_{\scriptscriptstyle \pm .002}$ & $0.26_{\scriptscriptstyle \pm .01}$ & $7.95_{\scriptscriptstyle \pm .03}$ & $11.26_{\scriptscriptstyle \pm .12}$ \\
\midrule

% --- COCO Image Captioning ---
 \textbf{Captioning}& $B@4$ $\uparrow$ & $CIDEr$ $\uparrow$ & $FD$  & $KD_{rbf}$  & $LMI$ & $MINE$ \\
\cmidrule(r){2-3} \cmidrule(lr){4-5} \cmidrule(l){6-7}
mPLUG & $39.5$ & $130.9$ & -- & -- & -- & -- \\
Hiera+neck & $36.1_{\scriptscriptstyle \pm .1}$ & $116.8_{\scriptscriptstyle \pm .1}$ & $68.9_{\scriptscriptstyle \pm .2}$ & $81.4_{\scriptscriptstyle \pm 1.6}$ & $6.37_{\scriptscriptstyle \pm .05}$ & $8.51_{\scriptscriptstyle \pm .01}$ \\
SAM2+neck & $35.4_{\scriptscriptstyle \pm .3}$ & $114.2_{\scriptscriptstyle \pm .9}$ & $71.1_{\scriptscriptstyle \pm .4}$ & $82.3_{\scriptscriptstyle \pm 1.1}$ & $6.35_{\scriptscriptstyle \pm .01}$ & $8.47_{\scriptscriptstyle \pm .06}$ \\
\bottomrule
\end{tabular}
}
\end{table}

This section analyzes the performance trade-offs between the Hiera and SAM2 encoders, examines the impact of knowledge distillation and neck capacity, and quantifies the representational bottlenecks of sequential adaptation via our cross-neck methodology.

\subsection{Overall Task Performance}

\cref{tab:performance_summary} highlights the performance trade-offs associated with model specialization. In monocular depth estimation, a task related to segmentation, the specialized SAM2 encoder (3.07 RMSE) outperforms its generalist predecessor, Hiera (3.21 RMSE), and achieves performance comparable to the task-specific DPT baseline. This confirms that specialization effectively refines features for spatially-aligned tasks.

However, this specialization creates an information bottleneck for other tasks. In pose estimation and image captioning, the generalist Hiera encoder is superior, outperforming SAM2 in AP (73.3 vs. 69.8) and CIDEr (116.8 vs. 114.2).
This suggests that the fine-tuning process for segmentation discards high-level information, crucial for complex tasks, thereby confirming the representational cost of specialization. 
Such a trade-off is fundamental in a universal feature coding context, even before accounting for a rate constraint. 

A deeper analysis of the feature metrics in \cref{tab:performance_summary} reveals a complex dynamic. For instance, in depth estimation, Hiera shows a slightly higher mutual information (10.4 LMI) with the baseline than SAM2 (10.3 LMI), despite its worse performance. 
This indicates that raw information content or feature distribution similarity alone do not dictate performance. We hypothesize that this is because performance depends on both the information content and feature transformability. SAM2's features, while less versatile, are more easily transformed by the neck for depth estimation. Hiera's features are richer and more general, making them superior for complex tasks like captioning, even if their raw distribution is more distant from the task-specific baseline.

\begin{table}[tb]
\caption{Effect of using distillation loss on the performance of the \texttt{neck\_6L} model across three tasks. Results shown using SAM2 encoder for Depth and Hiera for Pose and Caption.}
\begin{center}
\setlength{\tabcolsep}{3pt}

\begin{tabular}{ll|ccc}
\toprule
\textbf{Task} & \textbf{Training Scheme} & Performance & $FD$ $\downarrow$ & $LMI$ \\
\midrule

\multirow{2}{*}{Depth (RMSE $\downarrow$)}
% & No Distillation  & 3.22 & 291 & 7.6 \\
% & With Distillation & 3.08 & 12 & 9.6 \\
& No Distillation  & 3.21 & 231 & 7.7 \\
& With Distillation & 3.07 (+4.4\%) & 9.3 & 10.3 \\
\midrule

\multirow{2}{*}{Pose (AP $\uparrow$)} 
% & No Distillation & 70.1 & 1060 & 8.2 \\
% & With Distillation & 71.0 & 128 & 10.1 \\
& No Distillation & 72.1 & 1565 & 6.7 \\
& With Distillation & 73.3 (+1.7\%) & 0.3 & 8.1 \\
\midrule

\multirow{2}{*}{Caption (CIDEr $\uparrow$)} 
% & No Distillation & 98.4 & 102.6 & 5.1 \\
% & With Distillation & 115.3 & 40.9 & 6.2 \\
& No Distillation & 106.1 & 494 & 4.2 \\
& With Distillation & 116.8 (+10.1\%) & 68.9 & 6.4 \\

\bottomrule
\end{tabular}
\label{tab:training_schemes_restructured}
\end{center}
\end{table}

% \vspace{-0.2cm}
\subsection{Impact of Knowledge Distillation and Neck Capacity}

\textbf{Knowledge Distillation:}
Employing a distillation-based objective consistently and significantly improves performance across all tasks (\cref{tab:training_schemes_restructured}). To illustrate the maximum potential of distillation, we report its effect on the best-performing encoder for each task (SAM2 for the aligned depth task, and Hiera for pose and captioning tasks). In SAM2, depth estimation RMSE decreases from 3.21 to 3.07. When adapting the generalist Hiera, the benefits of distillation vary by task: for pose estimation, AP increases from 72.1 to 73.3, while the most substantial gain appears in image captioning, where CIDEr rises from 106.1 to 116.8 (+10.1\%). These results suggest that the distillation provides a crucial guiding signal, enabling the neck to transform Hiera’s features to better align with the specialized distributions expected by task decoders.

Feature-level metrics corroborate this: on all tasks, distillation drastically reduces Fréchet Distance (FD) and increases estimated mutual information (LMI). This confirms distillation is a powerful mechanism for closing the feature-space gap, aligning source representations with task-specific needs. Distillation acts as a powerful regularizer, guiding the neck's transformation to respect expert feature structures while minimizing task loss. It effectively bridges the ‘transformability gap’ between encoder features and task decoder expectations.

\begin{figure}[tb]
    \centering
    \subfloat[Depth Estimation\label{fig:depth_vs_neck}]{
        \includegraphics[width=0.3\linewidth]{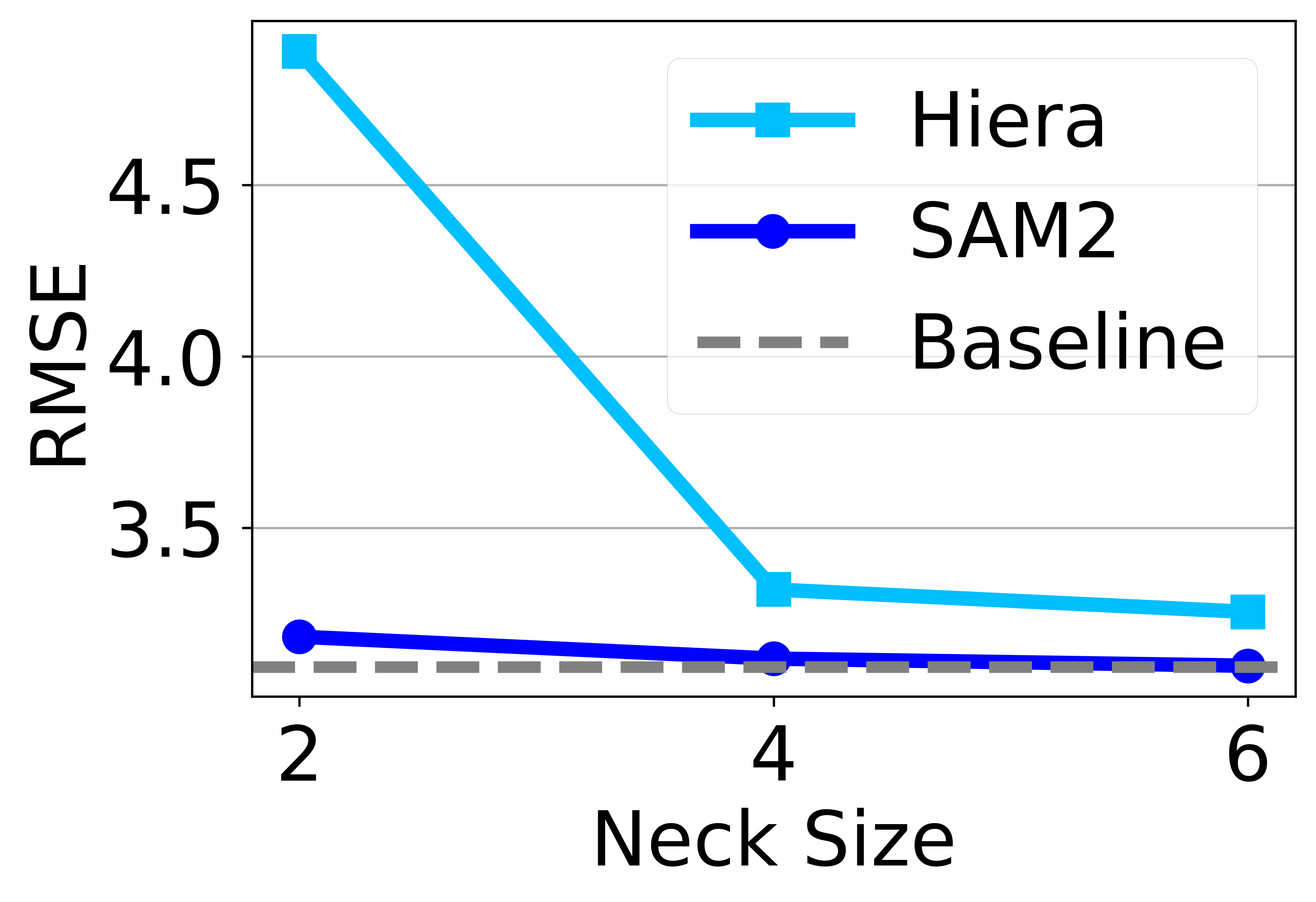}
    }
    \hfill
    \subfloat[Pose Estimation\label{fig:pose_vs_neck}]{
        \includegraphics[width=0.3\linewidth]{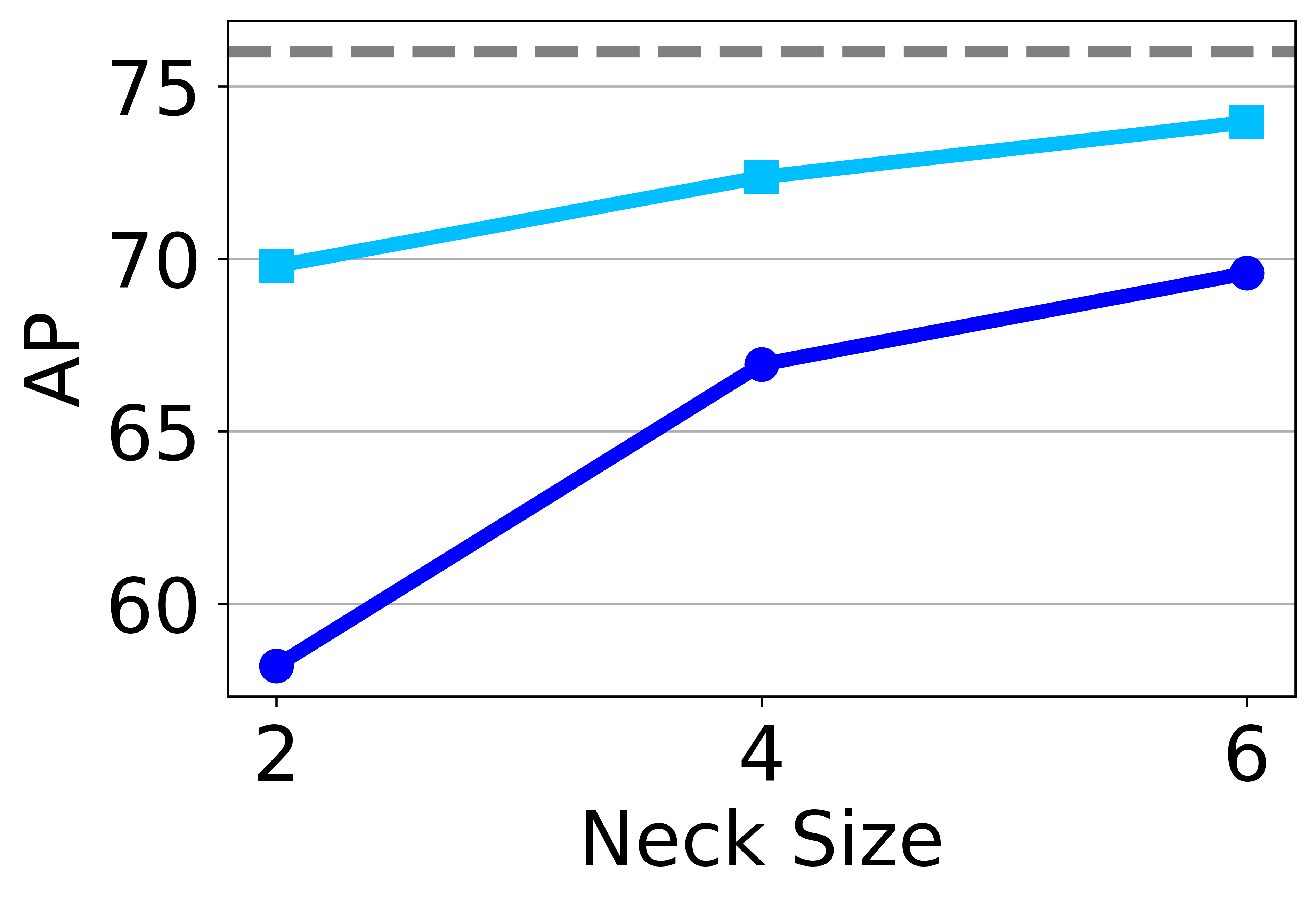}
    }
    \hfill \subfloat[Image Captioning\label{fig:caption_vs_neck}]{
        \includegraphics[width=0.3\linewidth]{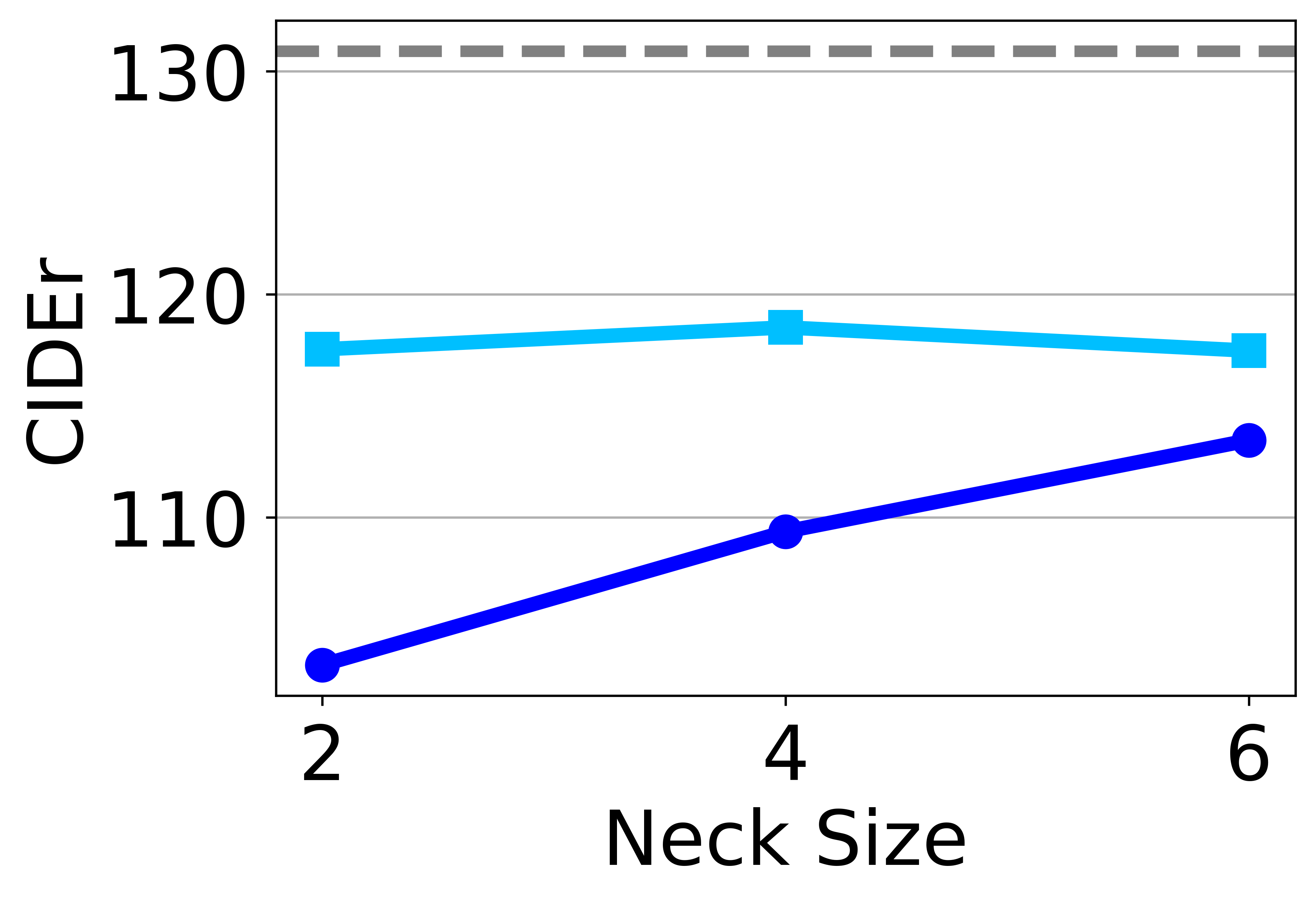}
    }
    \caption{The effect of Neck size on performance in distillation setting. Performance is measured in RMSE for Depth (lower is better), AP for Pose (higher is better), and CIDEr for Captioning (higher is better).}
 \label{fig:task_performance_vs_neck_size}
\end{figure}

\textbf{Neck Network Capacity:}
Neck capacity critically influences performance, but its impact depends on the encoder-task synergy (\cref{fig:task_performance_vs_neck_size}), revealing distinct behaviors for the generalist Hiera and specialist SAM2 encoders.
For depth estimation (\cref{fig:depth_vs_neck}), the specialized SAM2 gains little from a larger neck, as its features are already well-aligned. In contrast, Hiera's performance improves dramatically as neck size increases from 2 to 4 layers, showing that a larger neck is crucial for adapting its general-purpose features for a spatially-focused task.

Conversely, for pose and captioning, where Hiera excels, a larger neck improves SAM2's performance but fails to close the significant gap. This highlights that a capable neck can transform existing features but cannot recover information lost during an encoder's specialization.
Interestingly, the impact of neck capacity is highly dependent on the encoder-task pairing. For instance, while a larger neck provides substantial gains for Hiera on depth estimation, its effect on image captioning is minimal (\cref{fig:caption_vs_neck}). The performance for captioning shows little variation between the 2, 4, and 6-layer necks, indicating that %a shallow adapter is sufficient and 
increasing capacity offers no meaningful benefit. This suggests that for certain feature sets and tasks, additional adapter complexity does not necessarily improve performance and may not be a worthwhile trade-off. Therefore, selecting the ideal adapter capacity requires balancing the need for transformative power against the point of diminishing returns inherent to each unique encoder-task pairing.

\begin{figure}[tb]
    \centering

    % Row 1: Depth Estimation
    \subfloat[Feat. Dist. (Depth)\label{fig:depth_dist}]{
        \includegraphics[width=0.3\columnwidth]{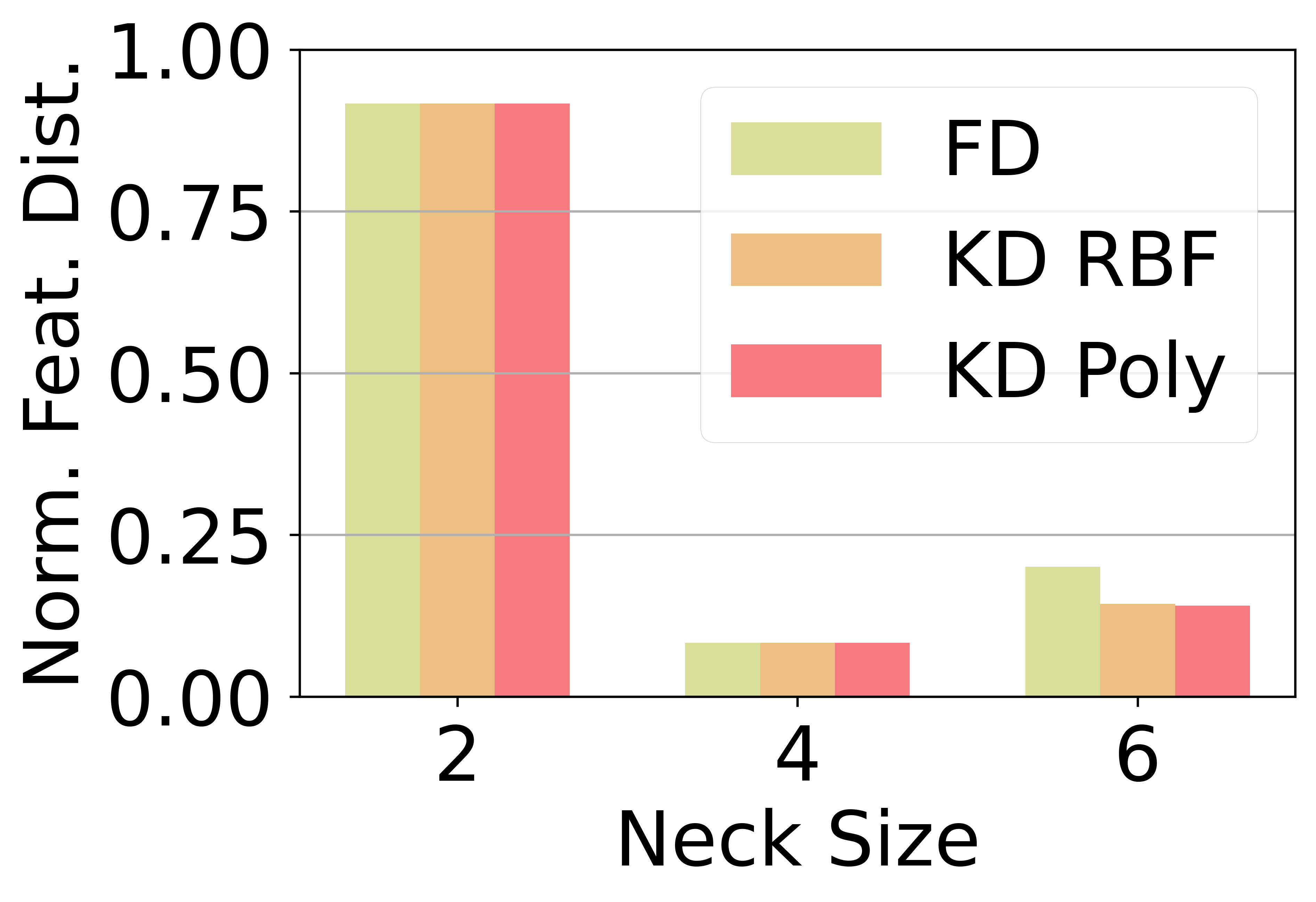}
    }
    \hfill 
    \subfloat[Feat. Sim. (Depth)\label{fig:depth_sim}]{
        \includegraphics[width=0.3\columnwidth]{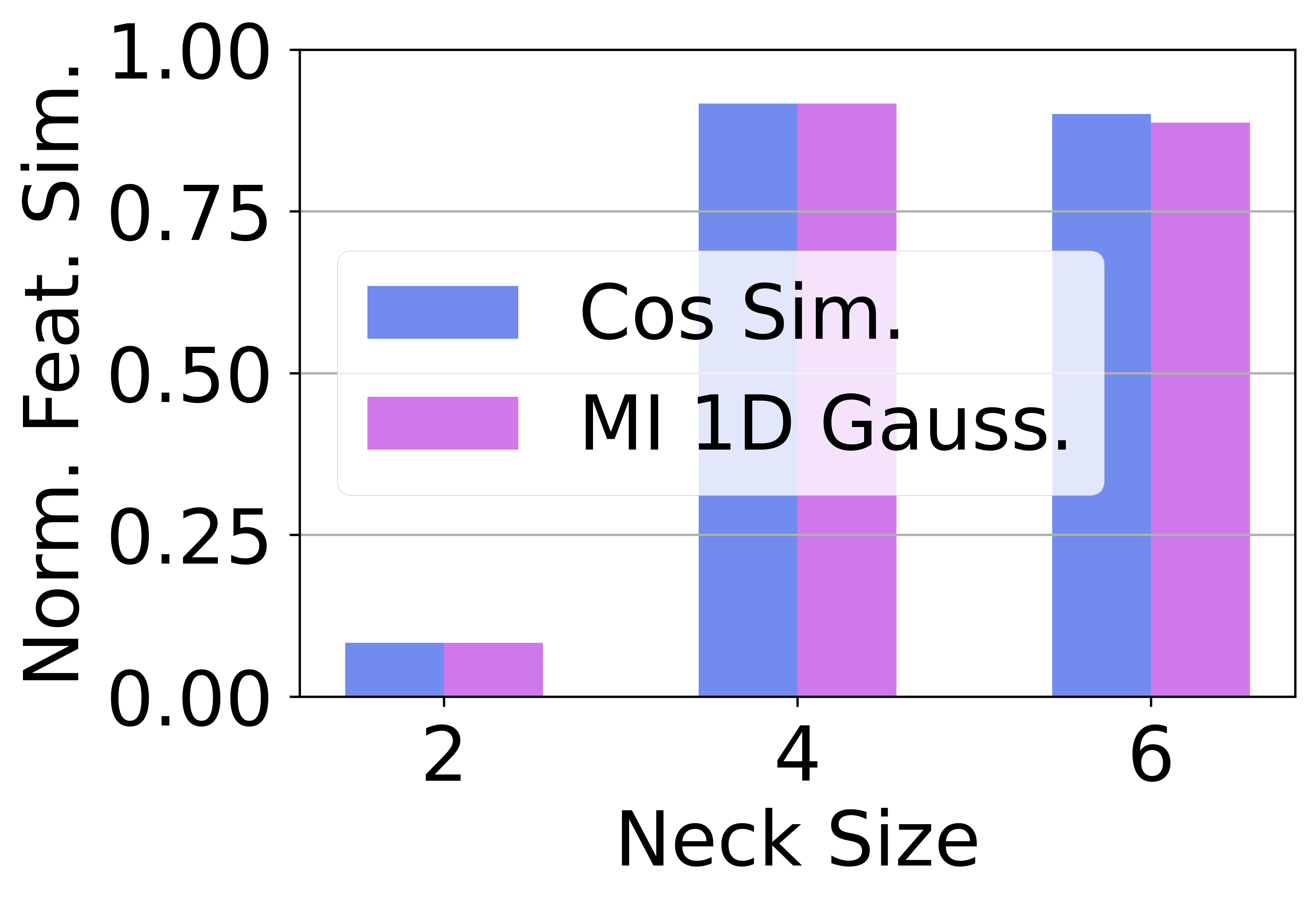}
    }
    \hfill
    \subfloat[MI Est. (Depth)\label{fig:depth_mi}]{
        \includegraphics[width=0.3\columnwidth]{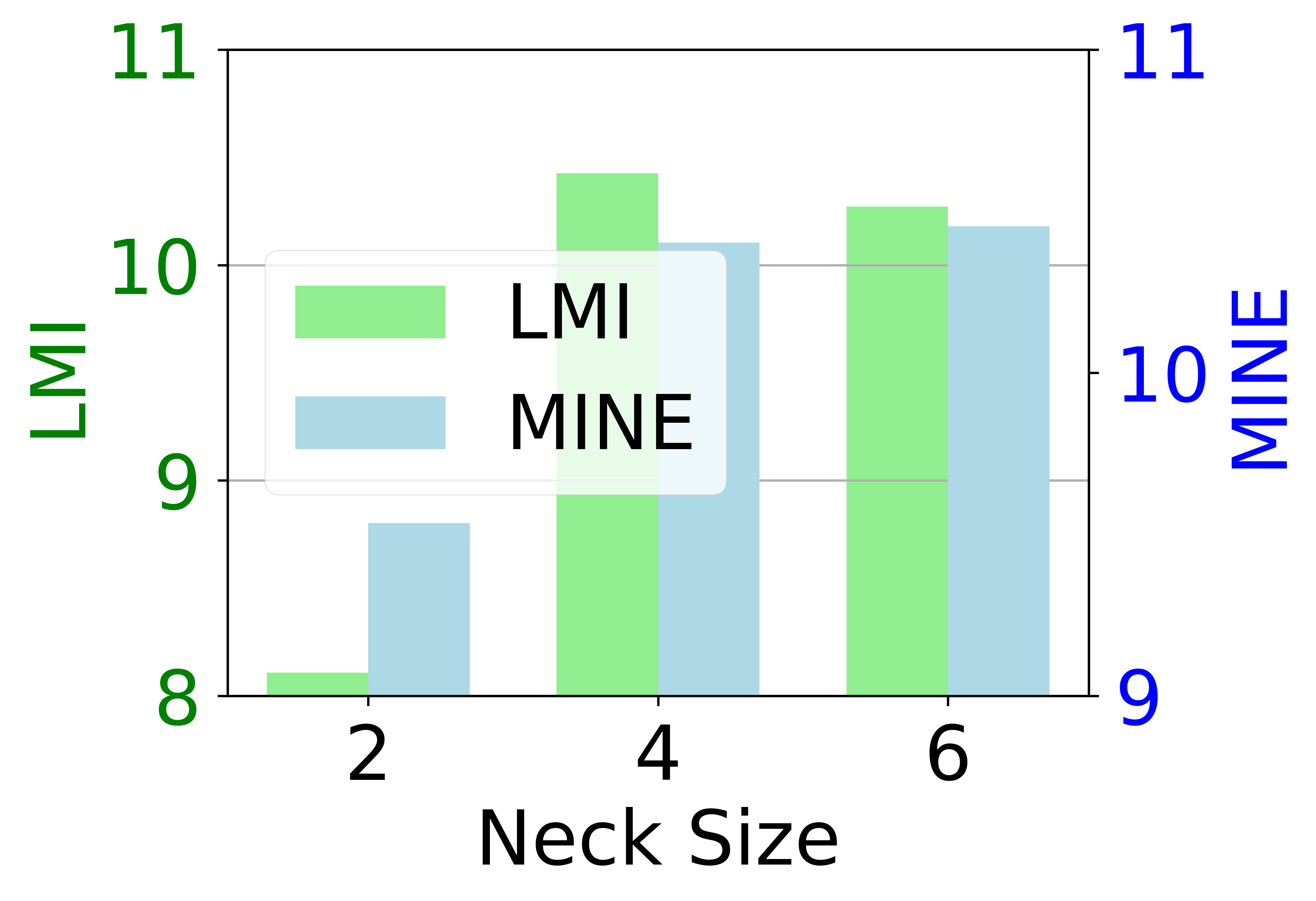}
    }
    
    % \vspace{-0.3cm} % Space between rows
    
    % Row 2: Pose Estimation
    \subfloat[Feat. Dist. (Pose)\label{fig:pose_dist}]{
        \includegraphics[width=0.3\columnwidth]{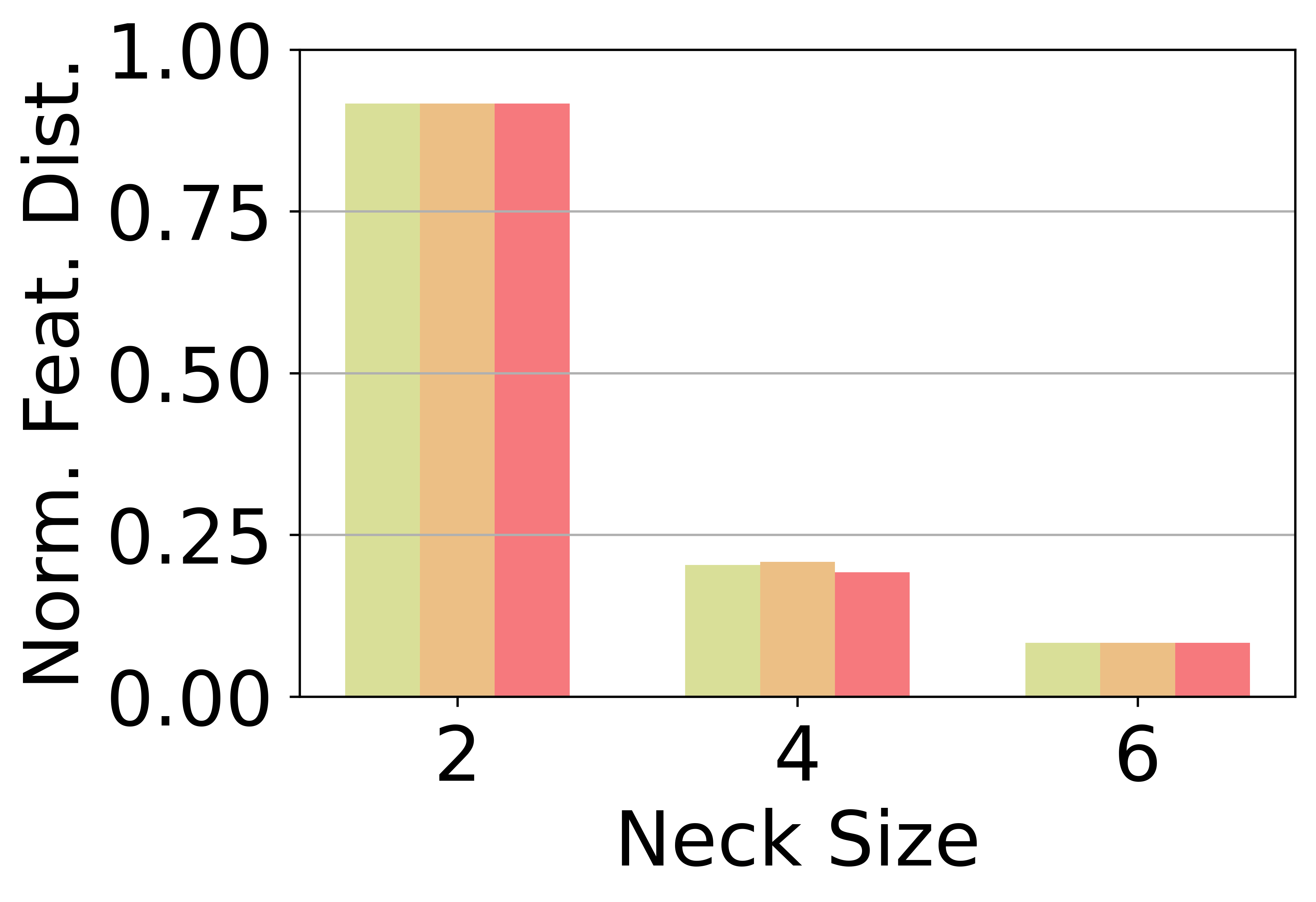}
    }
    \hfill 
    \subfloat[Feat. Sim. (Pose)\label{fig:pose_sim}]{
        \includegraphics[width=0.3\columnwidth]{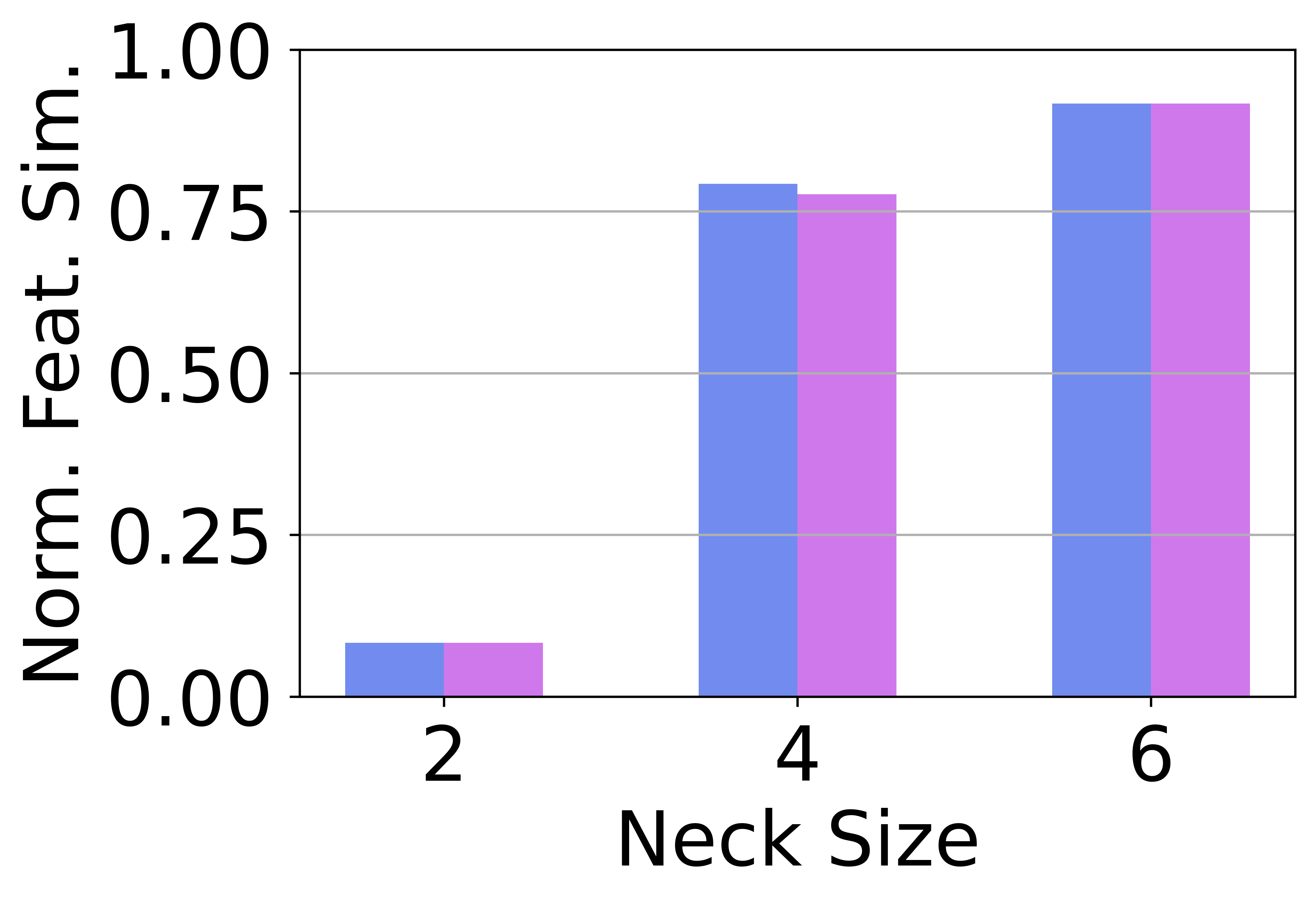}
    }
    \hfill
    \subfloat[MI Est. (Pose)\label{fig:pose_mi}]{
        \includegraphics[width=0.3\columnwidth]{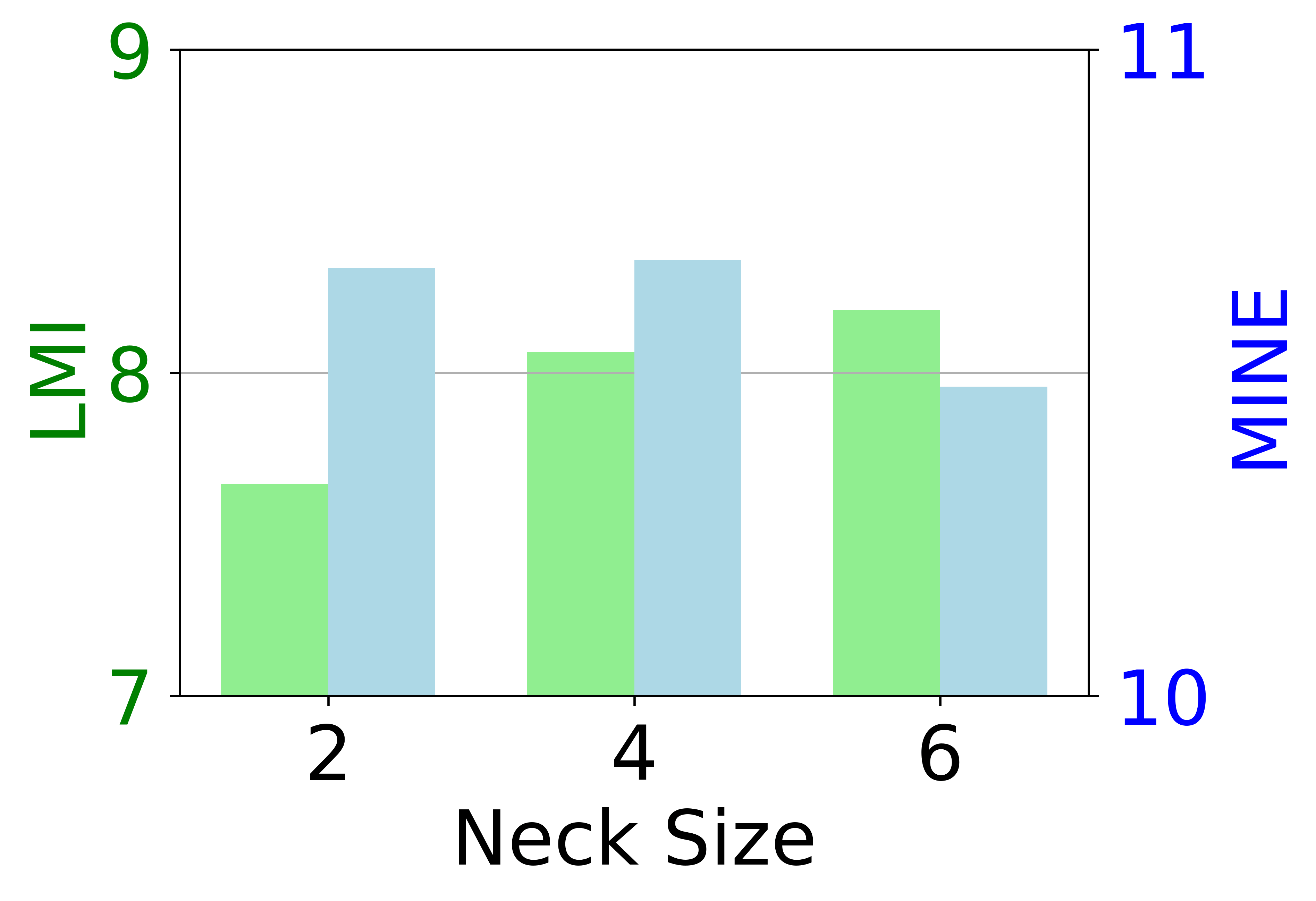}
    }

    % \vspace{-0.3cm} % Space between rows
    
    % Row 3: Image Captioning
    \subfloat[Feat. Dist. (Caption)\label{fig:caption_dist}]{
        \includegraphics[width=0.3\columnwidth]{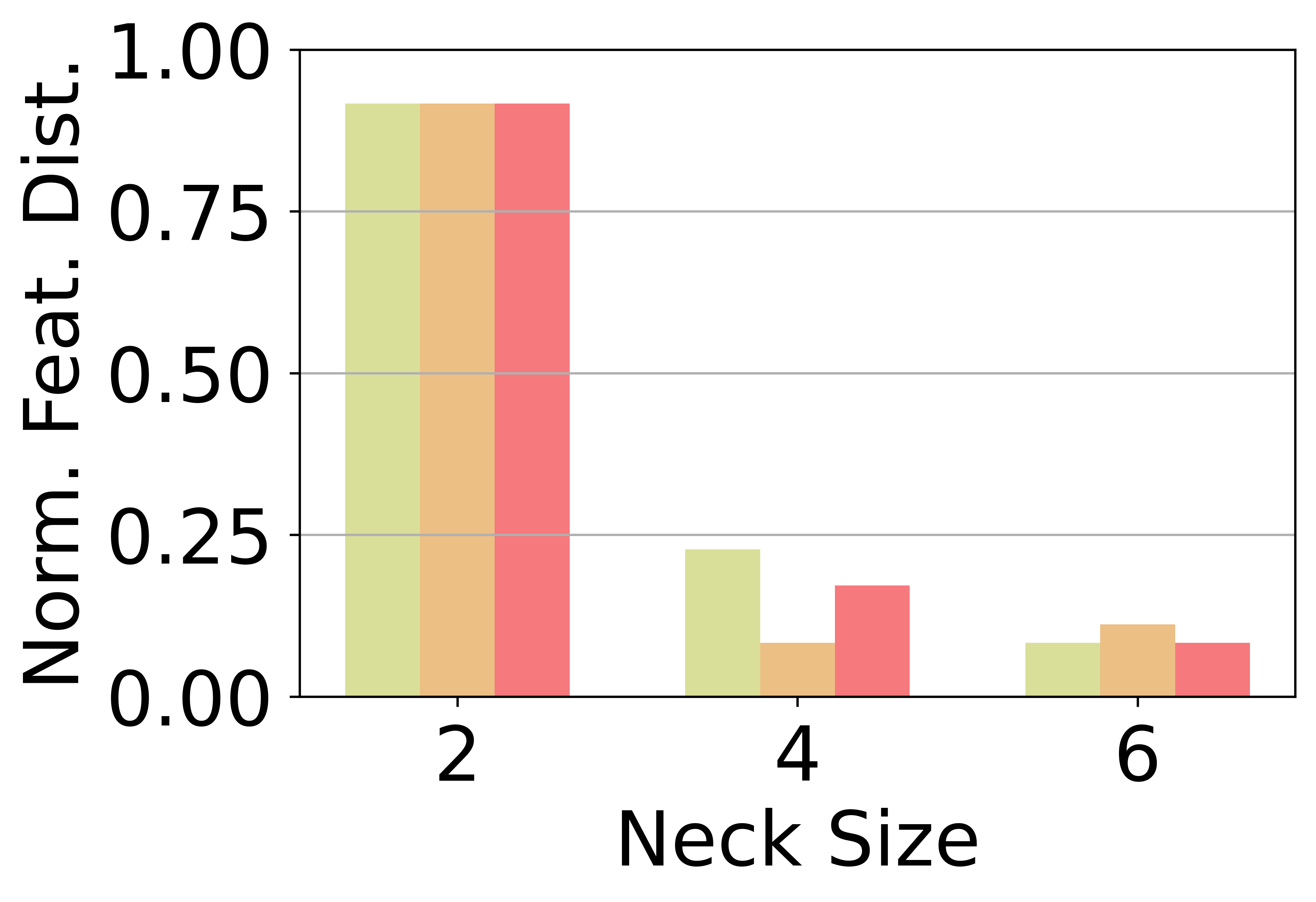}
    }
    \hfill 
    \subfloat[Feat. Sim. (Caption)\label{fig:caption_sim}]{
        \includegraphics[width=0.3\columnwidth]{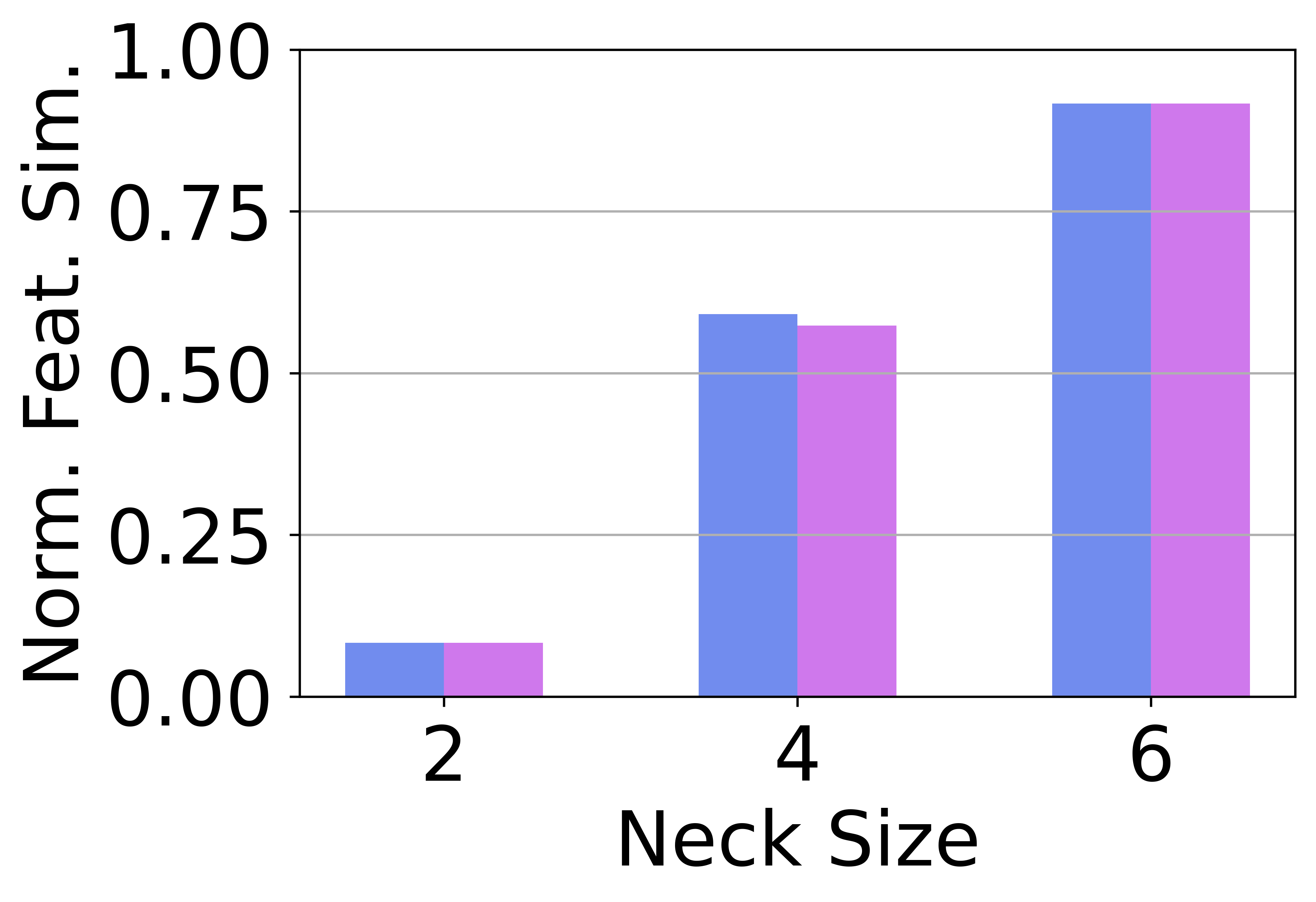}
    }
    \hfill
    \subfloat[MI Est. (Caption)\label{fig:caption_mi}]{
        \includegraphics[width=0.3\columnwidth]{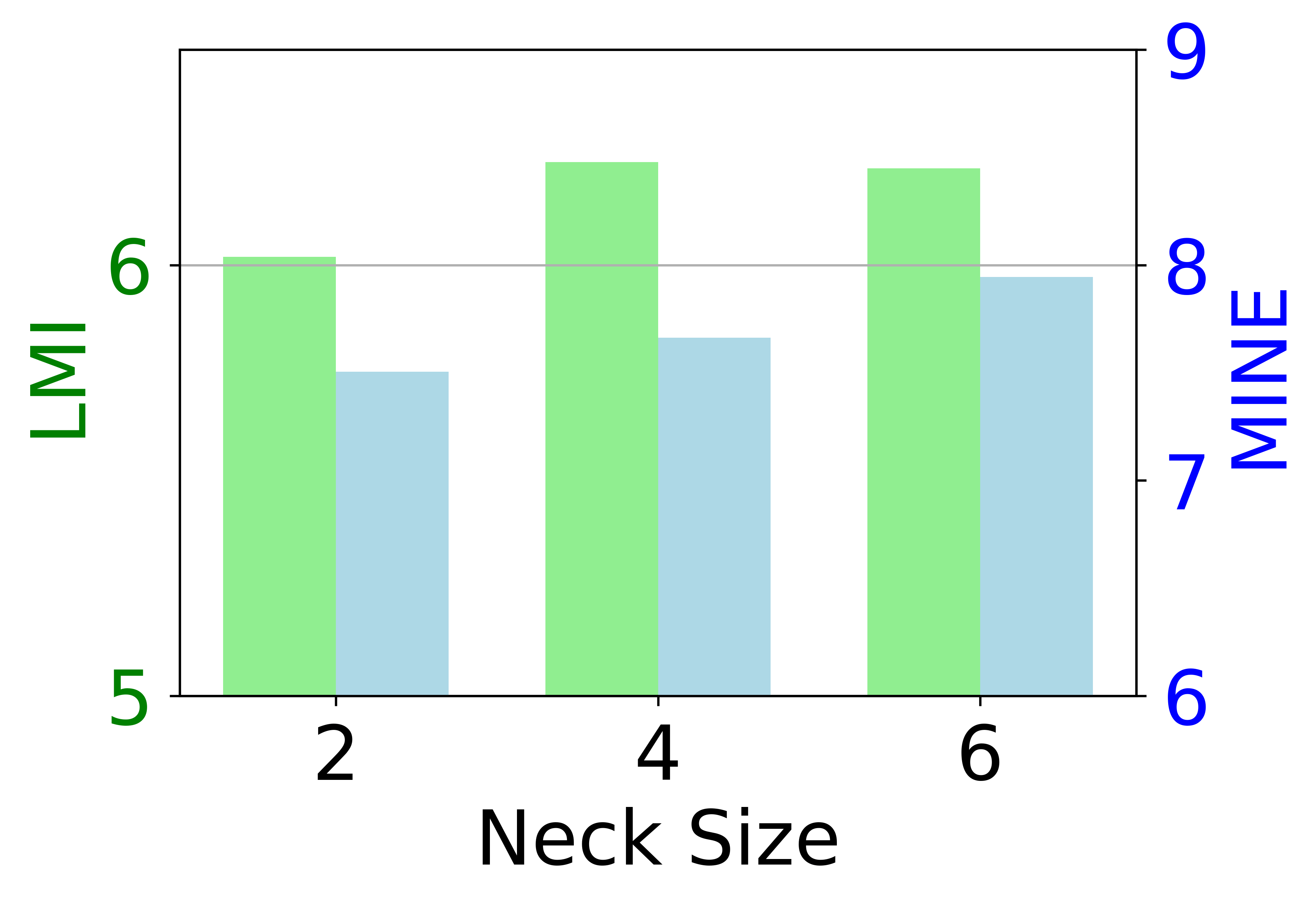}
    }

    \caption{
    Normalized feature distance/similarity, and mutual information (MI) vs. neck complexity for the SAM2 encoder across Depth Estimation (top),  Pose Estimation (middle), and Image Captioning  (bottom). For visualization purposes, each feature distance/similarity metric is independently scaled to its observed data range with 10\% padding at both ends to improve visual clarity.}
    \label{fig:all_tasks_feat_sim_dist_mi}
\end{figure}
% \vspace{-0.2cm}
\subsection{Feature-Level Information Analysis}

For the SAM2 encoder features, increasing neck capacity generally improves feature alignment (\cref{fig:all_tasks_feat_sim_dist_mi}), though the effect is most pronounced on tasks that are not yet performance-saturated. For pose and captioning, this trend is clear: feature distance metrics decrease while mutual information increases, supporting our hypothesis that a larger, distillation-guided neck better mimics an expert's feature distribution.

However, these metrics do not perfectly correlate with task performance. For depth estimation, where performance saturates quickly, feature metrics stabilize or show minor counter-intuitive shifts between 4- and 6-layer necks (\cref{fig:depth_dist,fig:depth_sim,fig:depth_mi}). This likely occurs because optimization finds functionally equivalent local minima with minor statistical differences once performance has plateaued. For pose estimation, the MINE estimate for the 6-layer neck drops slightly (\cref{fig:pose_mi}) even as LMI and performance improve, which may reflect MINE's known instability in high-dimensional spaces, making projection-based LMI more reliable.

Finally, for image captioning, LMI estimates (\cref{fig:caption_mi}) mirror the performance trend observed for Hiera, peaking at 4 layers and decreasing at 6. This reinforces the existence of an optimal neck capacity, beyond which a larger adapter may not improve information transfer, highlighting the complex relationship between network capacity, feature transformation, and final task performance. 
These nuances show that while feature metrics are insightful, they are best interpreted as part of a holistic analysis that includes end-to-end task performance.

\subsection{Cross-Task Adaptation and Representational Bottlenecks}

Using the \texttt{neck\_4L} for its strong balance of performance and efficiency, we employed our cross-neck method (\cref{fig:model_architecture_cross_neck}, \cref{sec:method}) to study how adapting features for one task creates a representational bottleneck for another.
Sequential adaptation invariably degrades final task performance compared to direct adaptation (\cref{tab:cross_neck_comaprison_with_baseline}). For instance, targeting depth after adapting for pose or captioning increases RMSE from 3.11 to 3.49 and 3.37, respectively. This shows that the process of adapting the features for the first task creates an irreversible information bottleneck, diminishing their value for any subsequent task.

The results reveal a strong trade-off with pose estimation. Adapting for pose first causes the most degradation on other tasks. Conversely, pose estimation is highly sensitive to prior adaptation, with its AP dropping from 66.9 to $\sim$58--60 when preceded by depth or captioning. This suggests pose estimation has distinct feature requirements, causing pronounced information loss when tasks are sequenced.

\begin{table}[tb]

\caption{Performance of the \texttt{SAM2+neck\_4L} model using direct vs cross-neck adaptation. Each neck in the sequence is denoted by the task for which it was optimized.}
\begin{center}
\begin{tabular}{ccccc}
\toprule
 \textbf{Neck Sequence} & Metric & Score &\textbf{$\Delta$ (\%)} & \textbf{LMI} \\
\midrule

% --- Evaluation on Depth Task ---
% & \textbf{RMSE} $\downarrow$ & \textbf{$\Delta$ (\%)} & \textbf{LMI} \\
% \cmidrule(lr){2-4}
% \multirow{3}{*}{Depth Estimation} 
\textbf{Depth} & \textbf{RMSE} $\downarrow$ &\textbf{3.11} & - & \textbf{10.4} \\
% \cmidrule(lr){1-2} \cmidrule(lr){3-5}
 Pose $\rightarrow$ Depth & & 3.49 & -12.2 & 9.4 \\
 Caption $\rightarrow$ Depth& & 3.37 & -8.4 & 9.5 \\
\midrule

% --- Evaluation on Pose Task ---
%  & \textbf{AP} $\uparrow$ & \textbf{$\Delta$ (\%)} & \textbf{LMI} \\
% \cmidrule(lr){2-4}
% \multirow{3}{*}{Pose Estimation} 
\textbf{Pose} & \textbf{AP} $\uparrow$ & \textbf{66.9} & - & \textbf{8.2} \\
% \cmidrule(lr){1-2} \cmidrule(lr){3-5}
Depth $\rightarrow$ Pose & & 58.1 & -13.2 & 7.7 \\
Caption $\rightarrow$ Pose & & 60.3 & -9.9 & 7.8 \\
\midrule

% --- Evaluation on Captioning Task ---
% & \textbf{CIDEr} $\uparrow$ & \textbf{$\Delta$ (\%)} & \textbf{LMI} \\
% \cmidrule(lr){2-4}
% \multirow{3}{*}{Image Captioning} 
\textbf{Caption} & \textbf{CIDEr} $\uparrow$& \textbf{109.3} & - & \textbf{6.2} \\
% \cmidrule(lr){1-2} \cmidrule(lr){3-5}
Depth $\rightarrow$ Caption & & 104.7 & -4.2 & 5.8 \\
Pose $\rightarrow$ Caption & & 97.6 & -10.7 & 5.7 \\

\bottomrule
\end{tabular}
\label{tab:cross_neck_comaprison_with_baseline}
\end{center}
\end{table}

\section{Conclusion}
\label{sec:conclusion}

Our direct comparison of the general-purpose Hiera encoder and its specialized descendant, SAM2, quantifies the information-theoretic cost of task specialization. The results reveal a clear trade-off: the specialist SAM2 excels on aligned tasks like depth estimation, while the generalist Hiera is superior for dissimilar tasks, such as pose estimation and image captioning. This confirms that specialization creates a representational bottleneck, enhancing performance on related tasks at the measurable cost of reduced versatility.

An important implication of our findings relates to universal feature coding, where a single encoder must extract information from images or video for a variety of downstream tasks. Our results demonstrate a core trade-off in information content between generalized and specialized feature extractors, even before a rate bottleneck is introduced. A specialized encoder offers distilled information and thus peak performance in a narrow domain, whereas a generalist provides greater overall utility for diverse applications. Furthermore, our cross-neck analysis demonstrated that this information loss is compounded by sequential adaptation, highlighting the fundamental compromise between specialization and generality.

We provide a quantitative foundation for designing information-aware adaptation strategies and for evaluating feature extractors. Though we focus on the Hiera-SAM2 lineage, our diagnostic framework can be used in future work to evaluate a variety of feature encoders, guiding the design of universal feature coding in emerging standardization efforts.

\bibliographystyle{IEEEtran}
\bibliography{IEEEabrv,references} %IEEEfull

\begin{thebibliography}{10}
\providecommand{\url}[1]{#1}
\csname url@samestyle\endcsname
\providecommand{\newblock}{\relax}
\providecommand{\bibinfo}[2]{#2}
\providecommand{\BIBentrySTDinterwordspacing}{\spaceskip=0pt\relax}
\providecommand{\BIBentryALTinterwordstretchfactor}{4}
\providecommand{\BIBentryALTinterwordspacing}{\spaceskip=\fontdimen2\font plus
\BIBentryALTinterwordstretchfactor\fontdimen3\font minus \fontdimen4\font\relax}
\providecommand{\BIBforeignlanguage}[2]{{%
\expandafter\ifx\csname l@#1\endcsname\relax
\typeout{** WARNING: IEEEtran.bst: No hyphenation pattern has been}%
\typeout{** loaded for the language `#1'. Using the pattern for}%
\typeout{** the default language instead.}%
\else
\language=\csname l@#1\endcsname
\fi
#2}}
\providecommand{\BIBdecl}{\relax}
\BIBdecl

\bibitem{dosovitskiy2021an}
\BIBentryALTinterwordspacing
A.~Dosovitskiy \emph{et~al.}, ``An image is worth 16x16 words: Transformers for image recognition at scale,'' in \emph{ICLR}, 2021. [Online]. Available: \url{https://openreview.net/forum?id=YicbFdNTTy}
\BIBentrySTDinterwordspacing

\bibitem{radford2021learning}
A.~Radford \emph{et~al.}, ``Learning transferable visual models from natural language supervision,'' in \emph{ICML}.\hskip 1em plus 0.5em minus 0.4em\relax PmLR, 2021, pp. 8748--8763.

\bibitem{oquab2024dinov}
\BIBentryALTinterwordspacing
M.~Oquab \emph{et~al.}, ``{DINO}v2: Learning robust visual features without supervision,'' \emph{TMLR}, 2024, featured Certification. [Online]. Available: \url{https://openreview.net/forum?id=a68SUt6zFt}
\BIBentrySTDinterwordspacing

\bibitem{ryali2023hiera}
C.~Ryali \emph{et~al.}, ``Hiera: A hierarchical vision transformer without the bells-and-whistles,'' in \emph{ICML}.\hskip 1em plus 0.5em minus 0.4em\relax PMLR, 2023, pp. 29\,441--29\,454.

\bibitem{ravi2025sam}
\BIBentryALTinterwordspacing
N.~Ravi \emph{et~al.}, ``{SAM} 2: Segment anything in images and videos,'' in \emph{ICLR}, 2025. [Online]. Available: \url{https://openreview.net/forum?id=Ha6RTeWMd0}
\BIBentrySTDinterwordspacing

\bibitem{chen2025samadapter}
\BIBentryALTinterwordspacing
T.~Chen \emph{et~al.}, ``{SAM}2-adapter: Evaluating \& adapting segment anything 2 in downstream tasks: Camouflage, shadow, medical image segmentation, and more,'' in \emph{ICLR 2025 Workshop on Foundation Models in the Wild}, 2025. [Online]. Available: \url{https://openreview.net/forum?id=iBoABx5wjA}
\BIBentrySTDinterwordspacing

\bibitem{sam_mpeg_fcm}
R.~Nguyen, C.~Rosewarne, and J.~Song, ``Fe2 related contribution: Introducing segment anything model,'' in \emph{{ISO/IEC JTC 1/SC 29/WG4, m73152}}, June 2025.

\bibitem{xu2022vitpose}
Y.~Xu, J.~Zhang, Q.~Zhang, and D.~Tao, ``Vitpose: Simple vision transformer baselines for human pose estimation,'' in \emph{NeurIPS}, vol.~35, 2022, pp. 38\,571--38\,584.

\bibitem{li2022mplug}
C.~Li \emph{et~al.}, ``mplug: Effective and efficient vision-language learning by cross-modal skip-connections,'' \emph{arXiv preprint arXiv:2205.12005}, 2022.

\bibitem{Geiger2013IJRR}
A.~Geiger, P.~Lenz, C.~Stiller, and R.~Urtasun, ``Vision meets robotics: The kitti dataset,'' \emph{Int. J. Robotics Research}, 2013.

\bibitem{eigen2015predicting}
D.~Eigen and R.~Fergus, ``Predicting depth, surface normals and semantic labels with a common multi-scale convolutional architecture,'' in \emph{Proc. IEEE/CVF ICCV}, 2015, pp. 2650--2658.

\bibitem{eigen2014depth}
D.~Eigen, C.~Puhrsch, and R.~Fergus, ``Depth map prediction from a single image using a multi-scale deep network,'' in \emph{NeurIPS}, vol.~27, 2014.

\bibitem{coco2014}
T.-Y. Lin \emph{et~al.}, ``Microsoft {COCO}: Common objects in context,'' in \emph{ECCV}, 2014, pp. 740--755.

\bibitem{chen2015microsoft}
X.~Chen \emph{et~al.}, ``Microsoft coco captions: Data collection and evaluation server,'' \emph{arXiv preprint arXiv:1504.00325}, 2015.

\bibitem{papineni2002bleu}
K.~Papineni, S.~Roukos, T.~Ward, and W.-J. Zhu, ``Bleu: a method for automatic evaluation of machine translation,'' in \emph{Proc. ACL}, 2002, pp. 311--318.

\bibitem{vedantam2015cider}
R.~Vedantam, C.~Lawrence~Zitnick, and D.~Parikh, ``Cider: Consensus-based image description evaluation,'' in \emph{Proc. IEEE/CVF CVPR}, 2015, pp. 4566--4575.

\bibitem{heusel2017gans}
M.~Heusel, H.~Ramsauer, T.~Unterthiner, B.~Nessler, and S.~Hochreiter, ``Gans trained by a two time-scale update rule converge to a local nash equilibrium,'' in \emph{NeurIPS}, vol.~30, 2017.

\bibitem{binkowski2018demystifying}
\BIBentryALTinterwordspacing
M.~Bi{\'n}kowski, D.~J. Sutherland, M.~Arbel, and A.~Gretton, ``Demystifying mmd gans,'' in \emph{ICLR}, 2018. [Online]. Available: \url{https://openreview.net/forum?id=r1lUOzWCW}
\BIBentrySTDinterwordspacing

\bibitem{belghazi2018mutual}
M.~I. Belghazi \emph{et~al.}, ``Mutual information neural estimation,'' in \emph{ICML}.\hskip 1em plus 0.5em minus 0.4em\relax PMLR, 2018, pp. 531--540.

\bibitem{gowri2024approximating}
G.~Gowri, X.~Lun, A.~Klein, and P.~Yin, ``Approximating mutual information of high-dimensional variables using learned representations,'' in \emph{NeurIPS}, vol.~37, 2024, pp. 132\,843--132\,875.

\end{thebibliography}

\end{document}